\documentclass[11pt,twoside,twocolumn,a4paper]{article}

\usepackage{cvwwcorr}
\usepackage{times}
\usepackage{epsfig}
\usepackage{graphicx}
\usepackage{amsmath}
\usepackage{amssymb}

\usepackage{bm}
\usepackage{sidecap}
\usepackage{floatrow}
\usepackage{cancel}
\usepackage{graphicx,color}
\graphicspath{{./fig/}}
\usepackage{xcolor}
\usepackage{enumitem}

\usepackage[pagebackref=true,breaklinks=true,bookmarks=false]{hyperref}
\cvwwfinalcopy 

\def\cvwwPaperID{5} 

\begin{document}

\title{On Learning Vehicle Detection in Satellite Video}

\author{Roman Pflugfelder\textsuperscript{1,2}, Axel Weissenfeld\textsuperscript{1}, Julian Wagner\textsuperscript{2}\\
\textsuperscript{1}AIT Austrian Institute of Technology, Center for Digital Safety \& Security\\
\textsuperscript{2}TU Wien, Institute of Visual Computing \& Human-Centered Technology\\
{\tt\small \{roman.pflugfelder|axel.weissenfeld\}@ait.ac.at, e1326108@student.tuwien.ac.at}
}

\maketitle
\ifcvwwfinal\thispagestyle{fancy}\fi

\begin{abstract}
Vehicle detection in aerial and satellite images is still challenging due to their tiny appearance in pixels compared to the overall size of remote sensing imagery. Classical methods of object detection very often fail in this scenario due to violation of implicit assumptions made such as rich texture, small to moderate ratios between image size and object size. Satellite video is a very new modality which introduces temporal consistency as inductive bias. Approaches for vehicle detection in satellite video use either background subtraction, frame differencing or subspace methods showing moderate performance (0.26 - 0.82 $F_1$ score). This work proposes to apply recent work on deep learning for wide-area motion imagery (WAMI) on satellite video. We show in a first approach comparable results (0.84 $F_1$) on Planet's SkySat-1 LasVegas video with room for further improvement.
\end{abstract}

\section{Introduction}
Object detection, i.e. the recognition and localisation of objects, in visual data is a very important and still unsolved problem. For example, the problem becomes challenging in aerial imaging and remote sensing as the data and scenes differ significantly from the case considered usually in computer vision~\cite{benenson-cvpr2013, ren-nips2015}.

Such remote detection is important in surveillance, as demanding applications let surveillance currently undergo a transition from near to mid distances (as with security cameras) to sceneries such as whole cities, traffic networks, forests, and green borders. Beside coverage new, low orbit satellite constellations\footnote{https://earthi.space, 11/03/2019} will allow multiple daily revisits and constantly falling costs per image. Such applications can be found e.g. in urban planning, traffic monitoring, driver behaviour analysis, and road verification for assisting both scene understanding and land use classification. Civilian and military security is another area to benefit with applications including military reconnaissance, detection of abnormal or dangerous behaviour, border protection, and surveillance of restricted areas.


\begin{figure}
\centering
\includegraphics[width=3cm]{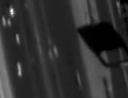}\vspace{1pt} \includegraphics[width=3cm]{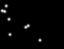}\\[1pt]
\includegraphics[width=3cm]{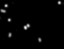}\vspace{1pt} \includegraphics[width=3cm]{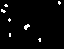}
\caption{Results of the proposed method. Top, left: video frame of the SkySat-1 LasVegas video showing a city highway with multiple cars. Top, right: vehicle labelling provided by Zhang et al.~\cite{zhang-rsens2019, zhang-corr2019}. Bottom, left: the method's response (heat) map. Bottom, right: the final segmentation result. The network detects all labelled cars and even a bus or truck at the right image border. \label{fig:title}}
\end{figure}

Although remotely acquired data shows great reduction of occlusion and perspective distortion due to the overhead view, new difficulties arise. Typical aerial and satellite images are very large in resolution and data size. For example, wide-area motion imagery (WAMI) provides instead of a few megapixel (MP) typical for security cameras up to 400\,MP per image frame and three image frames per second (2.2\,TB/s for 16\,bit per px). Satellite video gives today 4K RGB video with 30 frames per second (759.3\,MB/s). Satellite images capture large sceneries, usually dozens of square kilometers which introduce instead of a few visually large objects, thousands of tiny objects coming from hundreds of categories in a single image. At the same time these objects reduce in pixel size by orders of magnitude from $10^4$\,px to $10^2$\,px, to even $10$\,px for satellite video~\cite{zhang-rsens2019}, depending on the camera's ground sample distance (GSD)\footnote{GSD is the spatial distance of two adjacent pixels on the image measured on the ground.}.

This severe magnification of scenery and reduction of object size to very tiny appearances have consequences. Object detection becomes very ambiguous and sensitive to noise and nuisances and the search space dramatically increases and becomes very sparse. Inferred labels of data usually capture instead of the bounding box or contour sole positions, as the extent of objects is even for humans, e.g. in WAMI or satellite video, unrecognizable. All this leads to major difficulties if not inapplicability of vanilla methods~\cite{lalonde-cvpr2018}. Manual labelling of data is furthermore very tedious, for many cases impossible, hence, research on object detection in satellite video relies currently on background subtraction and frame differencing~\cite{ kopsiaftis-igarss2015, xu-jrsgis2017, yang-sensors2016, li-JARS2019, ao-corr2018, ao-tip2019}.

Recent literature~\cite{mou-igarss2016, mundhenk-eccv2016, zhang-spie2017, koga-rs2018, yang-corr2018, ding-jprs2018, guo-rs2018, lalonde-cvpr2018, zhang-rsens2019, imbert-thesis2019, AL-SHAKARJI_2019_CVPR_Workshops, Yang2019ClusteredOD, yang-iccv2019} also suggests to apply deep learning on aerial and satellite high resolution RGB single images, however, the work shows moderate performance for GSD larger than 15\,cm~\cite{mundhenk-eccv2016}. All work is also tested with rather narrow datasets of very different sceneries which makes the validity of the results questionable and the comparison of methods difficult. It is therefore unclear, if deep learning on high resolution images will further improve, given the limitations of the data.

Another problem of still images is the impossibility to capture the dynamic behaviour of vehicles which is essential for many applications. For example, vehicle heading and speed are important indicators in traffic models. Although rapid retargeting for multi-angular image sequences with Worldview-2 is possible~\cite{meng}, the time interval of around one minute between consecutive images is too large for reasonable analysis.


For these reasons the paper addresses the problem of vehicle detection in satellite video. Such video was introduced 1999 by DLR-TubSat, since 2013 Planet's SkySat-1 delivers up to 120\,s, 30\,Hz, 2K panchromatic video covering two areas of 1.1\,km\textsuperscript{2} with up to 80\,cm GSD. China's Jilin programme launched 2015, now provides even 4\,MP color video.


To the best of our knowledge this is the first work on using neural networks and deep learning to directly regress positions of vehicles \textit{in satellite video}. Inspired by recent work on WAMI~\cite{lalonde-cvpr2018} this paper proposes to exploit the temporal consistency in satellite video by using a neural network and deep learning instead of using background subtraction or frame differencing, by this improving over the state-of-the-art in vehicle detection with satellite video. To overcome shortage of labelled video, this work follows in this context the novel idea of transfer learning by recognising similarity of WAMI and satellite video data.

To summarise, the contributions of this work are
\begin{itemize}
\item the confirmation of results in LaLonde et al.~\cite{lalonde-cvpr2018} which shows clearly improvement in vehicle detection (from 0.79 to 0.93 in $F_1$ score) when using a spatiotemporal convolutional network,

\item empirical results showing the applicability of FoveaNet~\cite{lalonde-cvpr2018} to reduced resolution (0.91 $F_1$ score for 40\% of the original image resolution and 0.79 $F_1$ score for 20\%), yielding sizes of up to $3.6\times 1.8$\,px for vehicles which simulates satellite video and finally,

\item a transfer learning approach that uses labelled WAMI data to train a detector for satellite video with 0.84 $F_1$ score which is comparable to the currently best (subspace) method E-LSD\cite{zhang-corr2019} with 0.83 $F_1$ score on the same data.

\end{itemize}

\section{Related Work}
Deep learning significantly improved previously handcrafted methods of object recognition~\cite{benenson-cvpr2013}. Neural networks and back-propagation allow a learning formalism, where features and inference are jointly learnt from data in a neat end-to-end framework. Object detection is designed either as direct regression of bounding box image coordinates~\cite{redmon-corr2018} or by using the idea of object proposals as intermediate step~\cite{ren-nips2015}.

These developments triggered also work on deep learning for object detection in remote sensing~\cite{mou-igarss2016, mundhenk-eccv2016, zhang-spie2017, koga-rs2018, yang-corr2018, ding-jprs2018, guo-rs2018, lalonde-cvpr2018, zhang-rsens2019, imbert-thesis2019, AL-SHAKARJI_2019_CVPR_Workshops, Yang2019ClusteredOD, yang-iccv2019}. Applying deep learning for remote sensing is challenging, as labels are very expensive for satellite data and good augmentation, transfer learning or even unsupervised methods circumventing this problem are currently unknown~\cite{zhu-gsrsm2019, ma-isprs2019}.
Besides deep learning, object detection in remote sensing can be categorised according to the approach taken as well as the sensor modality, i.e. satellite image, sequence of multi-angular satellite images, satellite video, aerial image and WAMI.

Applying a classifier on top of a sliding window is one possible approach. Using a convolutional neural network in combination with hard negative mining showed by a $F_1$ score of 0.7 reasonable results with 15\,cm GSD on aerial images~\cite{koga-rs2018}. Following the golden standard~\cite{ren-nips2015}, adapted variants of the base feature, region proposal and Fast R-CNN network have been proposed such as using skip connections in the base and focal loss~\cite{yang-corr2018}, or using a dilated, multi-scale VGG16 as base in combination with hard negative mining~\cite{ding-jprs2018} which gives AP and Recall larger than 0.8 in their experiments. Guo et al.~\cite{guo-rs2018} introduces proprietary base, region proposal and detection networks, but did not show results on vehicles. This approach is useful with aerial images, but fails entirely for 1\.m GSD video as shown by~\cite{zhang-rsens2019} ($F_1$ score of 0.5). Results on high resolution satellite images are still unknown in literature.

Another idea is to pixel-wise classify vehicle vs. background (semantic segmentation), e.g. by combining Inception and ResNet to give a heatmap. Assuming a fixed vehicle size and using non-maxima suppression gives excellent results~\cite{mundhenk-eccv2016} ($F_1$ score larger than 0.9). Imbert proposes a generative U-Net in combination with hard negative mining for satellite images but kept unfortunately results in absolute $F_1$ scores confidential.

Spatiotemporal information is a further cue important in object detection, especially with WAMI and satellite video. The standard is to use background subtraction (BGS)~\cite{zhang-spie2017, kopsiaftis-igarss2015, xu-jrsgis2017, yang-sensors2016, ahmadi-ijrs2019} and frame differencing (FD)~\cite{li-JARS2019, ao-corr2018, ao-tip2019}, except Al-Shakarji et al.~\cite{AL-SHAKARJI_2019_CVPR_Workshops} who combined YOLO with spatiotemporal filtering on WAMI ($F_1$ score of 0.7), and Mou and Zhu~\cite{mou-igarss2016} who use KLT tracking on video with a SegNet on overlapping multispectral data, however, they did not show results for vehicles. Zhang and Xiang~\cite{zhang-spie2017} apply a ResNet classifier trained on CIFAR on proposals from a mixture of Gaussians foreground model, but did not show a proper evaluation.

The standard here is to apply connected component analysis~\cite{kopsiaftis-igarss2015, xu-jrsgis2017}, saliency analysis, segmentation~\cite{yang-sensors2016, li-JARS2019}, distribution fitting~\cite{ao-corr2018, ao-tip2019} followed by morphology. $F_1$ scores of larger than 0.9 for ships and scores between 0.6 and 0.8 for vehicles on the Burji Khalifa~\cite{yang-sensors2016}, Valencia~\cite{ao-corr2018, ao-tip2019} and Las Vegas~\cite{kopsiaftis-igarss2015} videos suggest BGS, FD for larger objects. Both BGS and FD depend heavily on registration and parallax correction, hence, these methods introduce various nuisances for vehicles which are difficult to handle. Evaluation on single, selective scenes is further too narrow to draw a final conclusion.

Very recent work~\cite{zhang-corr2019} suggests a subspace approach for discriminating vehicles and background. The idea shows potential with $F_1$ score results of larger than 0.8 on the simple Las Vegas video, which therefore needs further evaluation with more complex traffic patterns.

Another problem is the sparsity of vehicle occurrences in very large images as in WAMI which has been tackled by clustering the large images to draw attention to certain parts of the image and then to apply convolutional neural networks on single images~\cite{Yang2019ClusteredOD}\cite{yang-iccv2019} or multiple video frames~\cite{lalonde-cvpr2018} for final detection. Such clustering combined with deep spatiotemporal analysis shows excellent results on WAMI ($F_1$ score larger than 0.9)~\cite{lalonde-cvpr2018}.

Also very recently tracking of airplanes, trains and vehicles has been considered for satellite video~\cite{du-gsrsl2018, du-corr2018, shao-icme2018, guo-jstaeors2019}, either by using optical flow~\cite{du-gsrsl2018, du-corr2018}, correlation trackers (KLT)~\cite{shao-icme2018} or a combination of correlation and Kalman filters~\cite{guo-jstaeors2019}.

\begin{figure*}[t]
\begin{center}
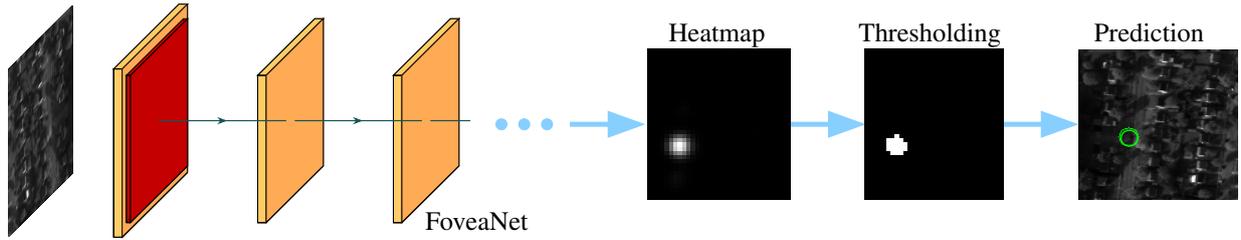 
\caption{The object detection process consists of two steps~\cite{lalonde-cvpr2018}: The FoveaNet predicts a heatmap, which indicates the likelihood that an object is at a given image coordinate. Vehicles are detected by thresholding the heatmap.}
\label{figure:overview}
\end{center}
\end{figure*}

\section{Methodology}
With our goal of detecting moving vehicles in satellite videos, we were inspired by the work of Lalonde et al. ~\cite{lalonde-cvpr2018}, who designed two neural networks, denoted as ClusterNet and FoveaNet, to detect vehicles in WAMI. The ClusterNet proposes regions of objects (ROOBI) based on areas of interest (AOI), which are input to the FoveaNet. Instead of using the ClusterNet to determine ROOBIs we split the AOI into square tiles (ROOBIs) with size $\medmuskip=0mu N \times N$; e.g. $N$=128\,px. The object detection based on the FoveaNet consists of two steps as depicted in Fig.~\ref{figure:overview}.

\subsection{FoveaNet and thresholding}

The FoveaNet is a fully convolutional neural network (CNN) and consists of eight convolutional layers. The number of filters per convolution are 32, 32, 32, 256, 512, 256, 256 and 1. Their filter sizes are summarized in Tab.~\ref{tab3}. After the first convolution a $\medmuskip=0mu 2 \times 2$ max pooling is carried out. Moreover, during training the 6$^{th}$ and 7$^{th}$ convolutional layers have a 50\% dropout. The heatmap is generated by the final $\medmuskip=0mu 1 \times 1$ convolutional layer where each neuron gives a vote of the likelihood of a moving vehicle at pixel level.

The input to the network is a stack of frames with size $\medmuskip=0mu N \times N \times c$, where $\medmuskip=0mu N \times N$ is the ROOBI size and $c$ depicts the number of consecutive adjoining frames in a stack. Hereinafter we refer to $c$ as channels. Thereby, the CNN shall learn to predict the positions of the objects of the central frame. We believe the FoveaNet is capable to learn spatiotemporal features by feeding the network with stacks of multiple frames (e.g. $c$=5), which are especially important in lower resolution images as existing in satellite videos.

The ground truth is based on heatmaps $H$, which are created by superimposing Gaussian distributions, where the center of each distribution is the pixel position ($x$,$y$) of the vehicle in the image:

\begin{equation}
H(x,y) = \sum_{n=1}^{N} \frac{1}{2 \pi \sigma^2} e^{-\frac{x^2+y^2}{2 \sigma^2}}
\label{eq:heatmap}
\end{equation}

where $n$ are the downsampled ground-truth coordinates provided in pixel positions and $\sigma$ is the
variance of the Gaussian blur. During training the network learns to minimize the Euclidean distance between the network output and the generated ground truth heatmaps.


The original FoveaNet uses ReLUs as activation functions. We discovered, however, the problem known as the “Dying ReLU” problem\footnote{http://cs231n.github.io/neural-networks-1, 11/03/2019}. During training, a weight update triggered by a large gradient flowing through a ReLU can make the neuron inactive. If this happens, the gradient flowing through this ReLU will always be zero and the network continues to give the same output. In our trainings we frequently discovered this phenomenon ($\sim$71\% of the cases) using the Xavier initialization ~\cite{glorot2010understanding}. Hence, we replaced the ReLUs with either ELUs (Exponential Linear Unit) or Leaky ReLUs.

\begin{figure*}[ht]
\begin{center}
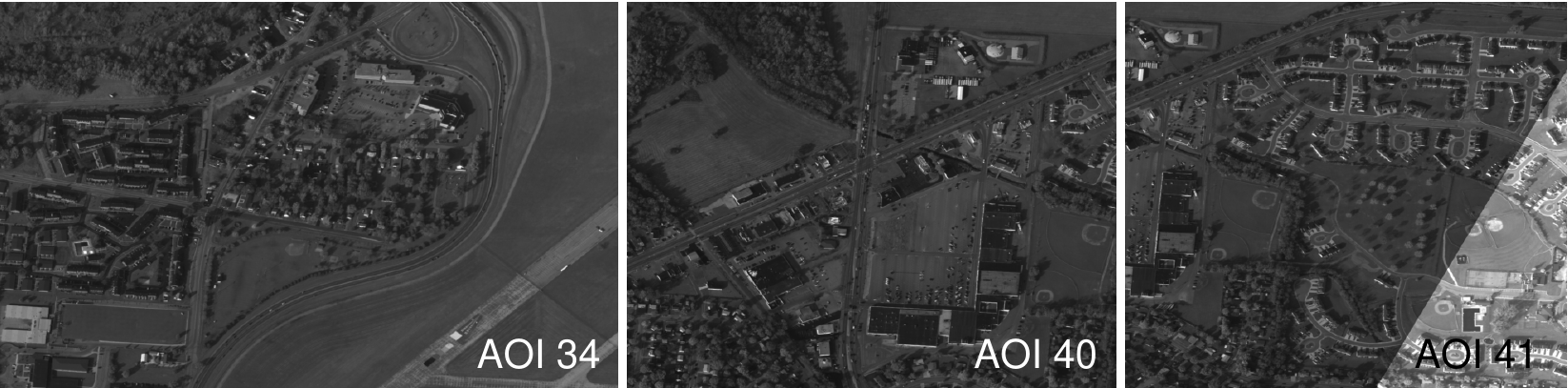 
\caption{AOI 40 contains a lot of dense traffic passing the intersection. On the contrary, AOI 41 contains mostly single vehicles driving on the road. Traffic patterns of AOI 34 are a combination of AOI 40 and AOI 41.}
\label{figure:aois_}
\end{center}
\end{figure*}


The second step processes the predicted heatmap to determine the objects' positions. For this, the heatmaps are converted into segmentation maps via OTSU thresholding~\cite{lalonde-cvpr2018}. If the segmented area is larger than a threshold $\alpha$, then the center of the area is defined as the object position.

\subsection{Transfer learning}
To the best of our knowledge there are currently no annotated datasets of satellite videos publicly available. In contrast, there are some labeled WAMI datasets accessible; e.g. the WPAFB dataset\footnote{https://www.sdms.afrl.af.mil/index.php?collection=wpafb2009, 11/03/2019} contains over 160.000 annotated moving vehicles. WAMI and satellite images, however, differ considerably, among other things due to the different GSD. For instance, the WPAFB images have about four times higher GSD than the LasVegas video. Our core idea is to use transfer learning for a domain transfer from WAMI to satellite images. For this, we train our CNN based on the WPAFB dataset. Afterwards we fine-tune the CNN on satellite video data.

\section{Experimental Evaluation and Results}
Our network was trained from scratch using PyTorch - we used Adam with a learning rate of 1e-5 and a batch size of 32. Data preparation includes frame registration to compensate camera motion.

We conducted three experiments. In the first experiment we carry out a baseline evaluation to reproduce the results of \cite{lalonde-cvpr2018}. For the second experiment, we reduce the image resolution (GSD) of the WPAFB dataset. Thereby, the vehicle size in these low-resolution images is in the same order as in satellite videos. In the third experiment, we carry out a fine-tuning and evaluate the FoveaNet on satellite data.

Detections are considered true positives if they are within a certain distance $\theta$ of a ground truth coordinate. If multiple detections are within this radius, the closest one is taken and the rest, if they do not have any other ground truth coordinates within the distance $\theta$, are marked as false positives (FP). Any detections that are not within $\theta$ of a ground truth coordinate are also marked as FP. Ground truth coordinates which have no detections within $\theta$ are marked as false negatives. Quantitative results are compared in terms of precision, recall, and $F_1$ measure. 

To compare our results with LaLonde et al.~\cite{lalonde-cvpr2018} we selected three of their AOIs (area of interest) - 34, 40 and 41. The contents of the AOIs 40 and 41 with respect to traffic patterns widely differ as displayed in Fig. \ref{figure:aois_}. Whereas AOI 40 contains a lot of dense traffic at a main intersection, AOI 41 mainly consists of single vehicles on the road. AOI 34 is a combination of both traffic patterns. Data was split into training and testing in the following manner: AOI 34 was trained on AOIs 40 and 41. AOI 40 was trained on AOIs 34 and 41 and AOI 41 was trained on 34 and 40. In contrast to~\cite{lalonde-cvpr2018}, we omitted AOI 42 for training as it is a sub-region of AOI 41.

For training and evaluation based on the WPAFB dataset, only frames with moving vehicles were included. We excluded frames without moving vehicles as our approach focuses solely on the detection and omits the region proposal part (ClusterNet) of \cite{lalonde-cvpr2018}. A vehicle is defined as moving if it moves at least $\omega$ pixel within 5 frames.

\begin{figure}[b]
\begin{center}
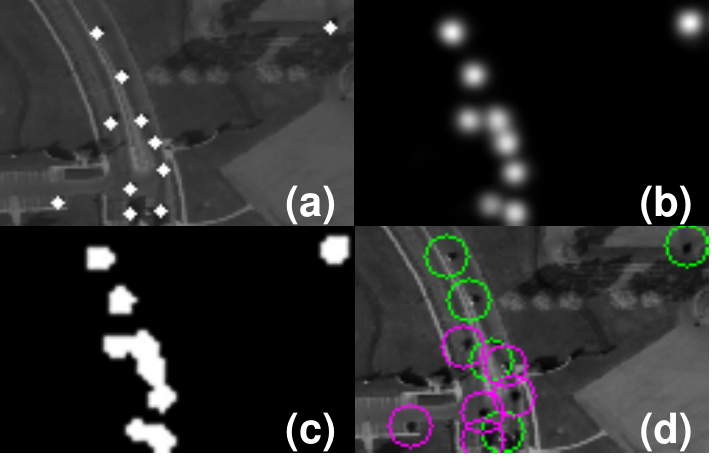 
\caption{The detection of vehicles in crowded scenes is error-prone. Detection results of a ROOBI with reduced resolution (SF=0.2): (a) ground truth, (b) predicted heatmap, (c) after thresholding, (d) detected vehicles (green: true positives, pink: false negatives)}
\label{figure:smallDistance}
\end{center}
\end{figure}

\subsection{Experiment 1: Baseline evaluation}
In the first experiment we reproduced the results in~\cite{lalonde-cvpr2018}. For this, we set the following parameters: $N$=100\,px (ROOBI edge length), $\sigma$=2 (variance of Gaussian blur), $\theta$=40\,px (evaluation threshold), $\omega$=15\,px (threshold for removing stationary cars) and $\alpha$=15\,px (threshold to disregard small segments). Tab.~\ref{tab1} indicates that our results are in the same order of magnitude than \cite{lalonde-cvpr2018}. For instance, we achieve a $F_1$ score of 0.90 in AOI 34 ($c$=5), whereas Lalonde et al. have a $F_1^*$ score of 0.93. The difference in the results is most likely due to the implementation differences of the second step, where we i.a. do not split connected regions into multiple detections. This presumption is confirmed looking at the evaluation results of AOI 40, where the differences of the $F_1$ score are greatest. AOI 40 contains a lot of dense traffic at the intersection resulting in connected regions, which cause false negative detections (Fig. \ref{figure:smallDistance}). Furthermore, the results in Tab.~\ref{tab1} confirm that the network is learning spatiotemporal features which improve the overall performance comparing single versus multi-channels. For instance, the precision of AOI 34 increases from 0.73 ($c$=1) to 0.87 ($c$=5).


\begin{SCtable*}
\small
\centering
\begin{tabular}{|c|c||c|c|c|c||c|c|c|c|c|c|}
\hline
\textbf{}&\textbf{}&\multicolumn{4}{c||}{\textbf{Experiment 1}} &\multicolumn{6}{c|}{\textbf{Experiment 2}}\\
\hline
\textbf{}&\textbf{}&\multicolumn{4}{c||}{\textbf{Full Resolution}} &\multicolumn{3}{c}{\textbf{Scaling factor 0.4}} &\multicolumn{3}{|c|}{\textbf{Scaling factor 0.2}}\\
\hline
\textbf{AOI} & \textbf{{\bm{$c$}}} & \textbf{Prec.}& \textbf{Rec.} & \bm{{$F_1$}} & \bm{{$F_1^*$}} & \textbf{Prec.}& \textbf{Rec.} & \bm{{$F_1$}} & \textbf{Prec.}& \textbf{Rec.} & \bm{{$F_1$}}\\
\hline
34 & 1 & 0.73 & 0.88  & 0.79 &  &  0.55	&0.55	&0.55 &  0.39	& 0.35	& 0.37\\
\hline
40 & 1 & 0.73 & 0.82 & 0.77 & & 0.55	& 0.48	& 0.51 & 0.20 &	0.21	& 0.20\\
\hline
41 & 1 & 0.76  & 0.90 & 0.82 & & 0.60	& 0.72	& 0.65 & 0.28 &	0.42 &	0.34\\
\hline
34 & 3 &  0.86 &	0.94	& 0.90 & & 0.93 &	0.77	& 0.84 & 0.80 &	0.61	& 0.69\\
\hline
40 & 3 & 0.92	& 0.89	& 0.90 & & 0.95	& 0.69	& 0.80 & 0.93	& 0.56 &	0.70\\
\hline
41 & 3 & 0.93	& 0.93 &	0.93 &  & 0.97	& 0.84	& 0.90 & 0.89	& 0.69	& 0.77\\
\hline
34 & 5 & 0.87 &	0.93 &	0.90& 0.93 & 0.94	& 0.78 &	0.85 & 0.91 &	0.63	& 0.74\\
\hline
40 & 5 & 0.92 &	0.89	& 0.90& 0.98 & 0.96 &	0.70	& 0.81 & 0.90 &	0.57 &	0.70\\
\hline
41 & 5 & 0.93 &	0.92	& 0.93 & 0.93 & 0.97 &	0.85	& 0.91 & 0.90	& 0.70	& 0.79\\
\hline
\end{tabular}
\caption{Results are based on three AOIs of the WPAFB dataset with various channel sizes ($c$). For comparison, $F_1^*$ scores of~\cite{lalonde-cvpr2018} are provided. Results of the second experiment include two scaling factors - 0.4 and 0.2. \label{tab1}}
\end{SCtable*}

\subsection{Experiment 2: Downscaled WPAFB dataset}
For the second experiment we reduced the images by a scaling factor (SF) of 0.4 and 0.2 resulting in 40\% and 20\% of the original image resolution, respectively. We selected a SF of 0.2, because this factor reduces the typical vehicle object size in the WPAFB dataset from the order of $\medmuskip=0mu 18 \times 9$\,px to $\medmuskip=0mu 3.6 \times 1.8$\,px, which is like the vehicle size in satellite videos. The following parameters were set for the experiments: SF=0.4 with $N$=100\,px, $\sigma$=2, $\theta$=16\,px, $\omega$=6\,px, $\alpha$=15\,px and SF=0.2 with $N$=100\,px, $\sigma$=1, $\theta$=8\,px, $\omega$=3\,px, $\alpha$=3.5\,px. Comparing results of detections based on $c$=1 (Tab. ~\ref{tab1}) indicate that the performance significantly decreases with lower image resolutions; e.g. the $F_1$ score of AOI 40 decreases from 0.77 to 0.20 (SF=0.2). In contrast, the detection results significantly improve if the number of channels is increased. These results confirm our hypothesis that the learned spatiotemporal features are of great importance for detecting tiny objects such as vehicles under low resolution.

\begin{figure}[b]
\centering
\includegraphics[width=8.0cm]{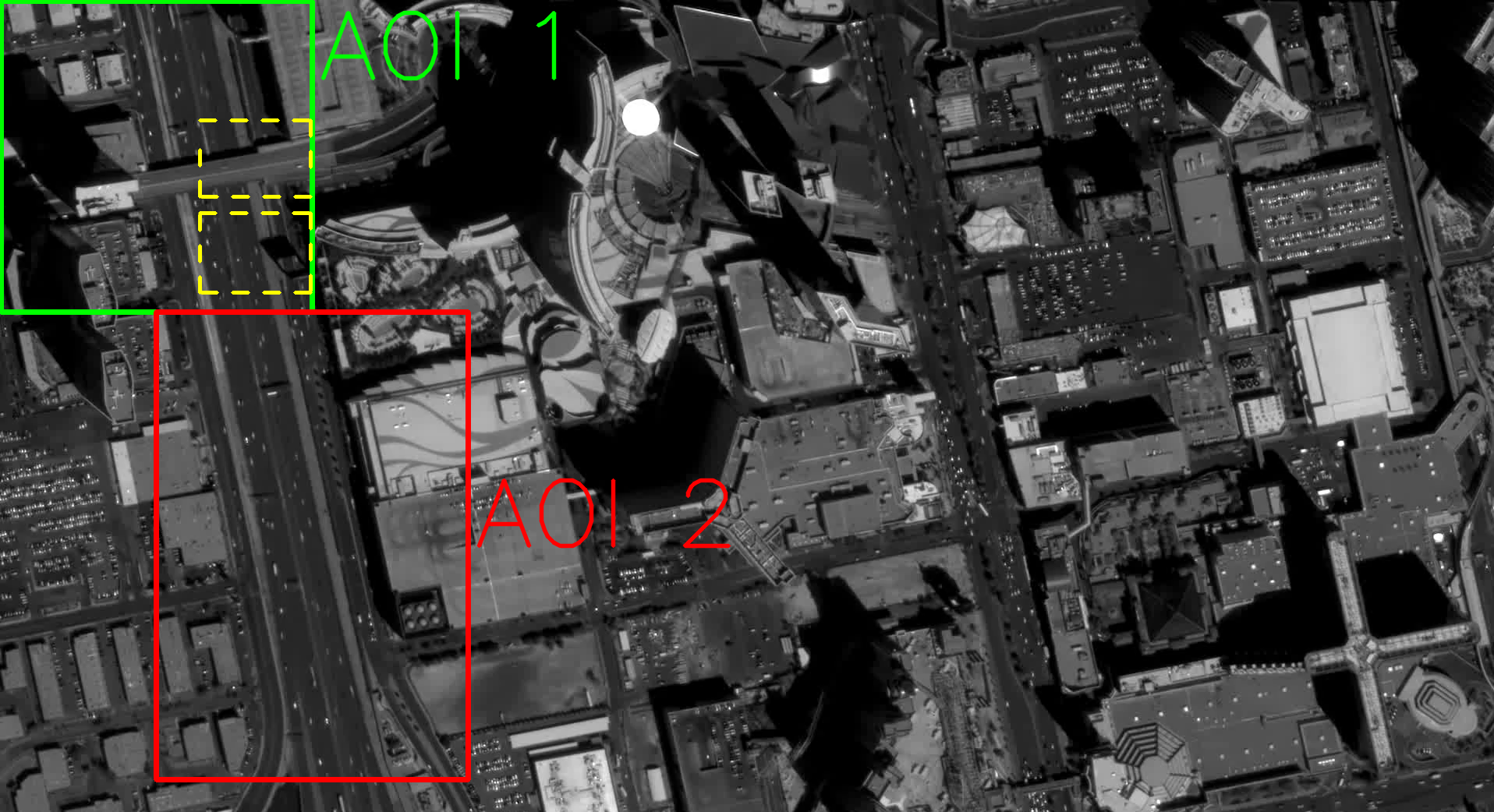}
\caption{Example of the SkySat-1 LasVegas video in which both AOIs are shown. AOI 1 (400x400\,px) is used for evaluation and AOI 2 (600x400\,px) for training. Two ROOBIs are sketched as yellow dashed rectangles.}
\label{fig:lasvegas}
\end{figure}





One of the main problems with low resolution images is the small distance between neighboring vehicles as displayed in Fig. \ref{figure:smallDistance}. In this case the FoveaNet creates a heatmap with a large number of connected regions, which result in a large number of false negative detections. To deal with small distances between neighboring cars we reduced the variance $\sigma$ of Eq. \ref{eq:heatmap}, which improved the detection results. Otherwise, this issue has not been addressed in this work, although enhancing step 2 of the object detection will most likely improve the results.


\subsection{Experiment 3: Satellite video}
The third experiment is conducted to evaluate the detection performance of the FoveaNet on the panchromatic satellite SkySat-1 LasVegas video\footnote{https://www.youtube.com/watch?v=lKNAY5ELUZY} consisting of 700 frames, whose GSD is $\sim$1.0\,m and its frame rate is 30\,fps. We defined two AOIs as illustrated in Fig.~\ref{fig:lasvegas}. While AOI 2 is mainly composed of straight parallel roads, AOI 1 contains additionally a bridge which results in more complex traffic patterns. The ground truth which was kindly shared by~\cite{zhang-corr2019} consists of bounding boxes for moving vehicles. We used the center points of those bounding boxes as ground truth analogous to the WPAFB ground truth.

For training and evaluation we set $\theta$=8\,px, $\alpha$=4\,px, $\sigma$=1, $c$=5 and $N$=128\,px. Additionally, we set SF=0.2 and $\omega$=3\,px for training the WPAFB dataset. We observed in this experiment higher efficiency in training by replacing the ELUs with Leaky ReLUs.

Tab.~\ref{tab2} shows the results of nine individual experiments using FoveaNet with different filter sizes in the respective convolutional layers (Tab.~\ref{tab3}). FoveaNet is trained on the 80\,\% reduced WPAFB and directly applied to the LasVegas video. We observe high recall ($>$0.8) but average precision which proves applicability of transfer learning.

In contrast to LaLonde et al.~\cite{lalonde-cvpr2018}, we do not observe large influence of the filter size to the final performance of the network. The argument that large filter sizes in the first layer are needed for spatial contextual information seems to be misleading, as context is introduced in higher layers of a deep network by the network's receptive field. We argue that the filter size depends on the pixel distance of vehicles in consecutive frames so that the spatiotemporal network can exploit temporal information which is empirically confirmed by our experiments.

We then choose slightly smaller filter sizes (13-11-9-7-5-3-3-1) for the convolutional layers in FoveaNet, as this configuration shows best final results. We fine-tuned the network on AOI 2 which improved $F_1$ score from 0.55 to 0.84. A qualitative result of this experiment is shown in Fig.~\ref{fig:title}. The heat map of the network reconstructs amazingly well the ground truth. It detects not only cars but also buses and trucks which the network never saw before. Three experiments with varying filter sizes show further that filter sizes have minor influence on the result. We clearly see that our proposed method outperforms most methods for vehicle detection in satellite video except E-LSD\cite{zhang-corr2019} which is comparable to our results.

\begin{figure}
\centering
\includegraphics[width=3cm]{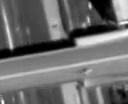}\vspace{1pt} \includegraphics[width=3cm]{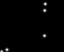}\\[1pt]
\includegraphics[width=3cm]{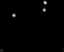}\vspace{1pt} \includegraphics[width=3cm]{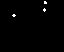}\\[1pt]
\includegraphics[width=3cm]{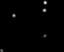}\vspace{1pt} \includegraphics[width=3cm]{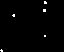}
\caption{From left to right. Top: input image and ground truth. Middle: estimated and thresholded heatmap, FoveaNet trained with AOI 2. Bottom: estimated and thresholded heatmap, FoveaNet after fine-tuning.\label{fig:1088}}
\end{figure}

We then performed an experiment where we directly trained all layers of FoveaNet on AOI 2. Surprisingly, the overall results are only slightly worse which indicates that the learning problem is not as complex as for the WPAFB dataset. We conclude from all observations that FoveaNet learns to detect moving spots by characterising the slope of linear movement in spacetime which is a much simpler learning problem as learning spatiotemporal changes of visual appearance. However, pre-training on WPAFB is important for the network to generalise as can be seen in Fig.~\ref{fig:1088}. Without pre-training the network is in this example not able to detect more complex motion patterns such as the moving vehicle on the bridge. It is an open question if such patterns could be learned by sole data augmentation.

Finally, we performed an experiment where we studied the effect of the frame rate of videos. Beside our baseline of considering every 10th image frame of the satellite video, we experimented with every 5th, 15th and 30th (1\,fps) video frame. The results indicate less influence of higher frame rates on performance. This again supports our hypothesis that very simple features such as typical slopes of vehicle trajectories in spacetime are learned by the network. This presumption needs however further experiments and insight.

\begin{table*}
\small
\caption{Left: Evaluation results of nine different filter size configurations (see Tab. \ref{tab3}) of the FoveaNet. Middle: Results of the FoveaNet trained from scratch, fine-tuned with different filter sizes and different fps (conf. 4). Right: Evaluation results of state-of-the-art (SOTA) methods are presented.}
\centering
\begin{tabular}{|c|c|c|c||c|c|c|c||r|c|c|c|}
\hline
\multicolumn{4}{|c||}{\textbf{WPAFB}} &\multicolumn{4}{c||}{\textbf{LasVegas AOI 1}} &\multicolumn{4}{|c|}{\textbf{SOTA}}\\
\hline
Conf. & \textbf{Prec.}& \textbf{Rec.} & \bm{{$F_1$}} &  & \textbf{Prec.}& \textbf{Rec.} & \bm{{$F_1$}} & & \textbf{Prec.}& \textbf{Rec.} & \bm{{$F_1$}}\\
\hline
1  & 0.56 & 0.67 & 0.61 & \text{scratch} & 0.87 & 0.80 & 0.83  & \text{ViBe\cite{barnich-tip2011}} & 0.58 & 0.17 & 0.26\\
\hline
2 & 0.46 & 0.76 & 0.57 & \multicolumn{4}{c||}{\textbf{fine-tuning}}  & \text{GMMv2\cite{zivkovic-icpr2004}} & 0.65 & 0.27 & 0.38\\
\hline
3 & 0.40  & 0.79 & 0.53  & 1 & 0.84 &0.82 & 0.83 & \text{GMM\cite{KaewTraKulPong2002}} & 0.46 & 0.50 & 0.48\\
\hline
\textbf{4} & \textbf{0.42} &	 \textbf{0.81} & \textbf{0.55} & \textbf{4} & \textbf{0.86} & \textbf{0.82} & \textbf{0.84} & \text{Fast-RCNN-LRP\cite{zhang-rsens2019}} & 0.58 & 0.44 & 0.50\\
\hline
5 & 0.43	& 0.85	& 0.58  & 9 & 0.76 & 0.85 & 0.80    & \text{GoDec\cite{zhou-icml2011}} & 0.95 & 0.36 & 0.52\\
\hline
6 & 0.47	& 0.80   & 0.60	& \text{skip} 5 & 0.84 & 0.83 & 0.84 & \text{RPCA-PCP\cite{candes-acm2011}} & 0.94 & 0.41 & 0.57\\
\hline
7 & 0.46 &	0.82 & 0.59 & \textbf{skip 10} & \textbf{0.86} & \textbf{0.82} & \textbf{0.84} & \text{Decolor\cite{zhou-tpami2013}} & 0.77 &	0.59	& 0.67\\
\hline
8 & 0.46 &	0.83	& 0.59 & \text{skip} 15 & 0.85 & 0.81	& 0.83 & \text{LSD\cite{liu-tip2015}} & 0.87 & 0.71 &	0.78\\
\hline
9 & 0.45 &	0.70	& 0.55  & \text{skip} 30 & 0.83 & 0.82 & 0.83 & \textbf{E-LSD\cite{zhang-corr2019}} & \textbf{0.85} & \textbf{0.79} & \textbf{0.82}\\
\hline
\end{tabular}
\label{tab2}
\end{table*}

\begin{table}
\small
\caption{Filter size configurations of the various experiments. Conf. 3 corresponds to the filter sizes suggested by LaLonde et al.~\cite{lalonde-cvpr2018}.}
\centering
\begin{tabular}{|c|l|c|l|}
\hline
\textbf{conf.} & \textbf{filter size} & \textbf{conf.} & \textbf{filter size}\\
\hline
1  & 19-17-15-13-11-9-7-1 & 6 & 9-7-5-3-3-3-3-1\\
\hline
2 & 17-15-13-11-9-7-5-1 & 7 & 7-5-3-3-3-3-3-1\\
\hline
3 & 15-13-11-9-7-5-3-1 & 8 & 5-3-3-3-3-3-3-1 \\
\hline
4 &13-11-9-7-5-3-3-1 & 9 & 3-3-3-3-3-3-3-1\\
\hline
5 & 11-9-7-5-3-3-3-1 & &\\
\hline
\end{tabular}
\label{tab3}
\end{table}
\section{Conclusion}
This paper considers vehicle detection in satellite video. Vehicle detection in remote sensing is challenging as the objects usually appear tiny compared to the size of typical aerial and satellite images and discrimination of objects of interest from background is frequently ambiguous. Satellite video is a very new modality introduced 2013 by Skybox (now Planet) which might overcome the problem by introducing high temporal resolution. This allows to exploit temporal consistency of moving vehicles as inductive bias. Current state-of-the-art methods use either background subtraction, frame differencing or subspace learning in video, however, performance is currently limited (0.26 - 0.82 $F_1$ score).

The method in this paper is motivated by recent work in WAMI which exploits video in spatiotemporal convolutional networks\cite{lalonde-cvpr2018}. We apply FoveaNet to the domain of satellite video by transfer learning the network with WPAFB and a small amount of available labelled video frames of the SkySat-1 LasVegas video which yields comparable results (0.84 $F_1$ score). Several ablation studies show minor influence of the filter sizes in the convolutional layers and minor influence of the frame rate (temporal resolution) on the overall result. This indicates a much simpler learning problem than for the original high-resolution WAMI data, however, we show that temporal information is essential for a good detection performance. Improvements of FoveaNet, e.g. including the final segmentation of the heat map into the network, are left for future work.

\section*{Acknowledgment}
This research was supported by the Austrian Research Promotion Agency (FFG) under grant MiTrAs-867030 and by the European Union's H2020 programme under grant FOLDOUT-787021. The imagery and video used in this work is in courtesy of the U.S. air force research laboratory sensors directorate layered sensing exploitation division and Planet Inc. under creative common  CC-BY-NC-SA\footnote{\url{https://creativecommons.org/licenses/by-nc-sa/4.0/legalcode}}.

{\small
\bibliographystyle{ieee}
\bibliography{literature}
}
\end{document}